\documentclass[11pt,a4paper,final]{article}
\usepackage[left=1in,right=1in,top=1in,bottom=1in]{geometry}
\usepackage{palatino} 
\usepackage{authblk}
\PassOptionsToPackage{sort&compress}{natbib}
\usepackage[numbers]{natbib}
\usepackage{booktabs} % For formal tables
\usepackage[ruled]{algorithm2e} % For algorithms
\usepackage{tikz}
\usetikzlibrary{arrows}
\usetikzlibrary{shapes.multipart}
\usepackage{caption}

\SetAlFnt{\small}
\SetAlCapFnt{\small}
\SetAlCapNameFnt{\small}
\SetAlCapHSkip{0pt}
\IncMargin{-\parindent}
\usepackage[most]{tcolorbox}
\usepackage[all]{nowidow}

\usepackage[font=small,labelfont=bf]{caption}
\usepackage[utf8]{inputenc} % allow utf-8 input
\usepackage[T1]{fontenc}    % use 8-bit T1 fonts
\usepackage{hyperref}       % hyperlinks
\hypersetup{
    colorlinks=false,
    linkcolor=blue,
    filecolor=magenta,      
    urlcolor=cyan,
    pdfpagemode=FullScreen,
    }
\usepackage{url}            % simple URL typesetting
\usepackage{booktabs}       % professional-quality tables
\usepackage{amsfonts}       % blackboard math symbols
\usepackage{nicefrac}       % compact symbols for 1/2, etc.
\usepackage{microtype}      % microtypography
\usepackage{xcolor}         % colors
\usepackage{amsmath}
\usepackage{mathtools}
\usepackage{amsthm}
\usepackage{amsmath}
\usepackage{amssymb}
\usepackage{bbm}
\usepackage{tikz} 
\usepackage{wrapfig}
\usepackage{floatrow}
\usepackage{enumitem}

\usepackage[suppress]{color-edits}
\usepackage{caption}
\usepackage{subcaption}

\usepackage{xcolor}

\title{Identifying, Evaluating, and Mitigating Risks of AI Thought Partnerships}

\begin{document}

\author[1]{Kerem Oktar*}
\author[2,3]{Katherine M. Collins*}
\author[2,3,4]{Jose Hernandez-Orallo}
\author[2]{Diane Coyle}
\author[2,3]{Stephen Cave}
\author[2,3,5]{Adrian Weller}
\author[6]{Ilia Sucholutsky}

\affil[1]{Princeton University}
\affil[2]{University of Cambridge}
\affil[3]{Leverhulme Centre for Future Intelligence}
\affil[4]{Universitat Politècnica de València,}
\affil[5]{The Alan Turing Institute}
\affil[6]{NYU}
\affil[*]{These authors contributed equally to this work.}
\date{}

\maketitle

\begin{abstract}
Artificial Intelligence (AI) systems have historically been used as tools that execute narrowly defined tasks. Yet recent advances in AI have unlocked possibilities for a new class of models that genuinely collaborate with humans in complex reasoning, from conceptualizing problems to brainstorming solutions. Such AI thought partners enable novel forms of collaboration and extended cognition, yet they also pose major risks—including and beyond risks of typical AI tools and agents. In this Commentary, we systematically identify risks of AI thought partners through a novel framework that identifies risks at multiple levels of analysis, including Real-time, Individual, and Societal risks arising from collaborative cognition (RISc). We leverage this framework to propose concrete metrics for risk evaluation, and finally suggest specific mitigation strategies for developers and policymakers. As AI thought partners continue to proliferate, these strategies can help prevent major harms and ensure that humans actively benefit from productive thought partnerships.
\end{abstract}

% \section{Introduction}

Artificial Intelligence (AI) systems have historically been used as tools that execute narrowly defined tasks, such as identifying faces in images. Yet recent advances in AI have unlocked possibilities for a new class of models that genuinely collaborate with humans in complex reasoning, from conceptualizing problems to brainstorming solutions; for instance, discussing the pros and cons of medical treatments with doctors, or planning military interventions with policymakers. Such AI thought partners \cite{collins_building_2024} or AITPs, enable novel forms of collaboration and extended cognition, yet they also pose major risks—including and beyond risks of typical AI tools and agents \cite{crockett_modern_2025}. Here, we propose a systematic framework for assessing risks of AITPs, offer metrics for the evaluation of these risks, and suggest potential strategies for risk mitigation.

There is a pressing need to understand these risks as AITPs move rapidly from development to deployment. New ‘reasoning’ models like Deepseek’s R1 or OpenAI’s o3, and new systems explicitly designed to collaborate in complex reasoning, such as Google’s ‘co-scientist’ \cite{gottweis_towards_2025} increasingly enable  impactful collaborative cognition—co-learning, co-creating, and co-planning—far beyond what is possible with other AI tools, including LLM agents \cite{shavit2023practices}. 

\section*{The RISc Framework}

Our RISc framework categorizes possible risks from collaborative cognition with AITPs into three interacting levels:  first, (R)eal-time risks arising from specific instances of human-AITP interactions; second, those arising from extended use by (I)ndividuals; and third, those arising when systems are deployed in (S)ociety in new kinds of collaborative (c)ognition. 
At each level, we separately consider risks inherent in an AITP’s performance (is the model performing appropriately?) and those that arise from how AITPs are utilized (is the model being used appropriately?), giving six risk categories (see Box 1).

\begin{tcolorbox}[
  colback=white,
  colframe=black,
  title={Box 1: Risk Categories},
  fonttitle=\bfseries,
  sharp corners,
  boxrule=0.5pt,
  breakable
]
\small
\noindent

Consider two impactful use cases for AITPs: medicine (a doctor uses an AITP to think through triage, diagnosis, and treatment planning) and policy (a policymaker uses an AITP to think through complex decisions during a financial crisis). We illustrate each risk category with one example from medicine and policy.
\noindent\rule{\linewidth}{0.4pt}
\vspace{0.5em}
\noindent\textbf{Real-time Risks}\par
\noindent\rule{\linewidth}{0.4pt}
\begin{description}[leftmargin=*, labelsep=1em, style=nextline]
  \item[Performance (generic deliberation)] Poorly-trained AITPs can encourage context-insensitive deliberation.  
    \emph{Medicine example:} A medical AITP may lead a doctor to consider globally prevalent but locally rare diseases, lowering efficiency of time-sensitive diagnosis.
  \item[Utilization (credit assignment)] Assigning praise and blame between users and AITPs is challenging when AITPs influence every stage of decisions.  
    \emph{Policy example:} Who is liable in a misdiagnosis if doctor and AITP deliberated together?
\end{description}
\noindent\rule{\linewidth}{0.4pt}

\vspace{0.5em}
\noindent\textbf{Individual Risks}\par
\noindent\rule{\linewidth}{0.4pt}
\begin{description}[leftmargin=*, labelsep=1em, style=nextline]
  \item[Performance (user manipulation)] An AITP can fail to benefit its user if it’s programmed to manipulate them.  
    \emph{Policy example:} An AITP may nudge a policymaker to author policies that benefit the AITP’s developer.
  \item[Utilization (cognitive atrophy)] Extensive use of AITPs can blunt critical reasoning skills.  
    \emph{Medicine example:} A doctor who always relies on an AITP may weaken their ability to form diagnostic hypotheses.
\end{description}
\noindent\rule{\linewidth}{0.4pt}

\vspace{0.5em}
\noindent\textbf{Societal Risks}\par
\noindent\rule{\linewidth}{0.4pt}
\begin{description}[leftmargin=*, labelsep=1em, style=nextline]
  \item[Performance (systemic fragility)] Centralization of AITP systems may induce fragility to blackouts or attacks.  
    \emph{Policy example:} Heavy reliance on AITPs in central banks could make monetary policy vulnerable to outages.
  \item[Utilization (homogeneity of thought)] Widespread AITP use may lead to convergence of opinion.  
    \emph{Policy example:} If all officials use the same AITP model, policies may fail to reflect local needs and knowledge.
\end{description}
\end{tcolorbox}

\section*{Evaluating Risks}
To systematically assess the presence and extent of these risks in genuinely collaborative cognition in practice, we need to  develop effective evaluations. Our framework both structures past work focused on assessing performance-level risks \cite{chan_harms_2023} and highlights new evaluations for utilization risks at multiple levels. 

\subsection*{Evaluating Real-time Risks}
Real-time risks depend on the thought interplay between users and AITPs and influence a community of stakeholders. For instance, consider an AITP helping a doctor decide which patients to admit to limited beds in a pandemic. A good AITP can help delineate and weigh trade-offs through an open dialogue; but how can we evaluate whether the tradeoffs proposed, and the dialogue process itself, are ‘good’? Such assessment requires considering the perspectives of multiple stakeholders—from the care team to the patients, their families, and the hospital and broader community writ large. 
The assessment of which tradeoffs are raised and how they are accounted for over the course of a conversation—over the thinking trace —could involve leveraging NLP methods or large language models to process the dialogue. Practitioners could specify what kinds of tradeoffs would be expected to be raised, which could be checked in the thinking trace; however, manual oversight of some form may be required—especially for evaluations that are not reducible to a single standardized “score,” such as the model’s moral judgment. 

These same measures of assessing the thinking trace itself could be used to evaluate the relative contributions of any human(s) and any AITP(s). For instance, if a medical researcher engages with an AITP in the construction of a clinical trial protocol, we could automatically tag a saved trace of the interactions to appraise both the amount and type of thinking that each thought partner contributed to the collective “product of thought.” 

\subsection*{Evaluating Individual Risks}
Such readouts could also be helpful in feeding back to the user, or supervisor of the user, to ensure that the user is contributing appropriately to the thought partnership and not over-relying on the model for ideas or insights.  But what does it mean for users to over- or under-rely on a model when the engagements unfold over multiple  interactions in this new kind of collaborative cognition? And how can we assess how user thinking may atrophy through over-reliance rather than developing through reasoned interaction? 
This may require regular intervals of assessment with a professional, with and without access to an AITP. Such assessments should draw inspiration from existing work evaluating team dynamics and other measures of team efficacy when one or more group members is removed for some stage in the collaborative reasoning process \cite{riedl_quantifying_2021}.  

Additionally, AITPs may require evaluation that looks at the user’s first-person experience in the reasoning process, rather than just the final output of collaborative thinking. Understanding how people interact with AITP in new forms of collaborative cognition requires interactive evaluation that engages with multiple steps in the reasoning process \cite{collins_evaluating_2024}. This could involve having the user give regular reports as to how they feel the interaction is progressing, which are later subject to qualitative and quantitative analysis. 
The expansion of AITPs in increasingly advanced realms of knowledge work, however, may increase the level of expertise required for external auditors in evaluation, especially in the space of natural language where outputs may look fluent but be incorrect. Financial incentives could be used to encourage experts who use AITPs to contribute to their development by providing feedback on their interactions, which can be used in further training.
\subsection*{Evaluating Societal Risks}
Additionally, evaluation of the risks of thought partnerships requires measures that assess group-level or societal risks. Consider the impact on homogeneity of thought: how can we know whether our collective thinking and intellectual output has begun to grow more homogeneous? We encourage the construction of ongoing metrics that measure thought diversity or any narrowing in research or other activities that could, for instance, track patents or other research output \cite{youn_invention_2015}. 
Such evaluation also requires deep engagement across fields: economists have measures for societal productivity but these lack engagement with how cognitive processes shape ideas or decisions and hence economic activity. All of these evaluations, however, hinge on more societal conversations around what we want from these systems. How do we weigh gains in productivity, creative expression, or joy—when we may now be thinking with machines? 
It is also important to note potential inequities in societal AITP deployment: the people who get access to such models first are likely to be those who already have access to good human collaborators (for instance, those in leading universities or companies). People who may benefit most from AITPs—those who currently lack access to mentors or advisors—could fall further “behind.”

\section*{Mitigations}

We next propose a series of potential mitigations to attenuate key risks. 

\subsection*{Mitigating Real-Time Risks}
At the instantial level, a lot of the effort in LLM research has focused on improving performance, but for AITP we need access-modulating strategies for different stages in the thinking process that can prompt the system to stop or delegate to humans as soon as  it is expected to fail. We can also imagine cases (such as recruitment) where we may restrict access to any thought partner, or ask for a record of the engagement in thought (for instance, all queries to an AITP and corresponding engagement). 
One mitigation that will be important is upskilling judgement. %Recent work has highlighted the role of researchers’ ability to judge whether or not they benefit from access to the AI tool. 
Courses teaching not just ideation, but critical judgement \cite{fischbach_problem_2024} will therefore be needed in a world of AITPs. 
Decentralizing, personalising and detailed logging of AITP systems (extending, for instance, the Retrieval-Augmented Generation \cite{zhou_metacognitive_2024} approach to metacognition, reasoning strategies, etc.) would mean that credit and responsibility could be better delineated in terms of intellectual property and the particular “engine” of thought used.

\subsection*{Mitigating Individual Risks}
At the individual level, a number of measures can reduce the risk of developers manipulating  users who may trust and rely on their AITP for critical decision-making. These include education about what situations  AITPs are well-suited for, about the level of control and incentives of the developers of the AI, as well as promoting competition and diversity among AI systems. Such education is especially important in the context of self-explanation, as being able to explain topics to oneself is key to learning \cite{lombrozo_learning_2024}. Building in regular practices of solo thinking regarding one’s own engagement of AITPs is likely to grow ever more important to ensure that thought partnerships enhance deep understanding, not merely illusions of understanding \cite{messeri_artificial_2024}, and that thought partners are genuinely aligned with users’ representations and intentions, rather than merely converging with their judgments \cite{oktar_dimensions_2024}.  

\subsection*{Mitigating Societal Risks}

Finally, at the societal level, we urge the promotion of competition amongst AITP developers, to reduce the risk of  widespread usage of a single AI system creating a monoculture of homogenous thought. One potential for mitigating fragility to blackouts or security lapses from widespread adoption of AITPs is to build more communities and spaces for human thought partnerships. Continuing to cultivate spaces for humans to engage, discuss ideas, plans and questions will help ensure that the new forms of collaborative cognition  remain sufficiently grounded to the world and each other, and help us continue to develop our own capacity for collective deliberation. 

We emphasize that these are only some possible mitigations. We encourage more conversation around other mitigations, and continuing reflection on whether chosen mitigations appropriately adapt to the rapidly evolving demands of AITPs.

\subsection*{Trade-Offs} 
Any decisions about measures and mitigations come with tradeoffs. For example, good evaluation may require collecting more data on professionals; however, this may come at the cost of privacy, both to the individual and potentially the organization they work for and its users. 

Similarly, measuring and guarding against homogeneity of thought risks paradoxically controlling thought. If we were just considering a standard large language model, we may imagine simply modulating the temperature (variability) of the output. However, for an AITP, we may want to structurally vary the kinds of outputs and forms of engagement in a thinking process (e.g., shifting to play a “devil’s advocate” rather than “Socratic partner” in deliberation). 

Thus, mitigating the risks of AITPs is not simply a task of implementing a fixed set of recommendations. Instead, domain experts and key stakeholders must jointly evaluate the trade-offs involved in the various mitigations to craft, implement, and monitor mitigations.

\section*{Conclusion}
Scholars have discussed the burdens and blessings of technologies that extend cognition for millenia \cite{plato_phaedrus_nodate}. The arrival of AITPs marks a turning point in this discussion: beyond tools that supplement specific capabilities (such as writing, which extends memory), AITPs can shape every aspect of reasoning—from ideation to planning—across many domains. Such transformative potential begets the responsibility to understand, evaluate, and mitigate risks. Our RISc framework facilitates this task by categorizing risks at three levels of analysis (the real-time, individual, and societal levels) into distinct classes (performance and utilization risks). 
Our analysis highlights a pressing need for interdisciplinary research on how to evaluate and mitigate utilization risks, and we highlight several concrete suggestions at each level. Our suggestions range from NLP-based evaluations of thinking traces to developing protocols that delegate reasoning to humans to balance the trade-off of performance and cognitive atrophy. These strategies can help humans develop genuinely beneficial thought partnerships with AI, and benefit from a historically unprecedented form of collaborative cognition. 

\section*{Acknowledgments}

We thank Umang Bhatt, Lujain Ibrahim, Kartik Chandra, and Lio Wong for valuable conversations that informed this work. KMC acknowledges support from King's College Cambridge and the Cambridge Trust. AW  acknowledges  support  from  a  Turing  AI  Fellowship  under grant  EP/V025279/1 and the Leverhulme Trust via CFI. This work is supported (in part) by ELSA - European Lighthouse on Secure and Safe AI funded by the European Union under grant agreement No. 101070617. Views and opinions expressed are however those of the author(s) only and do not necessarily reflect those of the European Union or European Commission. 

% Citations
% 1. Collins, K.M. et al, Nature Human Behavior 8 (10), 1851-1863 (2024).
% 2.Crockett, M.J., Nature Machine Intelligence (2025): 1-2.
% 3.Gottweis et al., Research Paper, Google, February (2025)
% 4.Shavit, Y. et al, Research Paper, OpenAI, December (2023).
% 5.Chan, A., et al,. in Proc. 2023 ACM Conf. Fairness Account. Transp. (ACM, 2023), pp. 651–666.
% 6.C. Riedl, et al., Proc. Natl. Acad. Sci. 118, e2005737118 (2021).
% 7.Collins, K. M., et al., Proc. Natl. Acad. Sci. 121, e2318124121 (2024).
% 8.Youn, H., Strumsky, D., Bettencourt, L. M., & Lobo, J,. J. R. Soc. Interface, 12(106), 20150272.(2015). 
% 9.A. Toner-Rodgers, arXiv:2412.17866 [econ.GN] (2024).
% 10. M. A. Fischbach, Cell 187, 1828 (2024).
% 11. Y. Zhou et al., in Proc. ACM Web Conf. (ACM, 2024), pp. 1453–1463. 
% 12. T. Lombrozo, Trends Cogn. Sci. 28, 1011 (2024).
% 13. L. Messeri, M. J. Crockett, Nature 627, 49 (2024). 
% 14. K. Oktar et al., Decision 11,4 (2024).
% 15. Plato, Phaedrus.

\bibliography{main}

\begin{thebibliography}{14}
\providecommand{\natexlab}[1]{#1}
\providecommand{\url}[1]{\texttt{#1}}
\expandafter\ifx\csname urlstyle\endcsname\relax
  \providecommand{\doi}[1]{doi: #1}\else
  \providecommand{\doi}{doi: \begingroup \urlstyle{rm}\Url}\fi

\bibitem[Chan et~al.(2023)Chan, Salganik, Markelius, Pang, Rajkumar, Krasheninnikov, Langosco, He, Duan, Carroll, Lin, Mayhew, Collins, Molamohammadi, Burden, Zhao, Rismani, Voudouris, Bhatt, Weller, Krueger, and Maharaj]{chan_harms_2023}
A.~Chan, R.~Salganik, A.~Markelius, C.~Pang, N.~Rajkumar, D.~Krasheninnikov, L.~Langosco, Z.~He, Y.~Duan, M.~Carroll, M.~Lin, A.~Mayhew, K.~Collins, M.~Molamohammadi, J.~Burden, W.~Zhao, S.~Rismani, K.~Voudouris, U.~Bhatt, A.~Weller, D.~Krueger, and T.~Maharaj.
\newblock Harms from {Increasingly} {Agentic} {Algorithmic} {Systems}.
\newblock In \emph{Proceedings of the 2023 {ACM} {Conference} on {Fairness}, {Accountability}, and {Transparency}}, {FAccT} '23, pages 651--666, New York, NY, USA, June 2023. Association for Computing Machinery.
\newblock ISBN 9798400701924.
\newblock \doi{10.1145/3593013.3594033}.
\newblock URL \url{https://dl.acm.org/doi/10.1145/3593013.3594033}.

\bibitem[Collins et~al.(2024{\natexlab{a}})Collins, Jiang, Frieder, Wong, Zilka, Bhatt, Lukasiewicz, Wu, Tenenbaum, Hart, Gowers, Li, Weller, and Jamnik]{collins_evaluating_2024}
K.~M. Collins, A.~Q. Jiang, S.~Frieder, L.~Wong, M.~Zilka, U.~Bhatt, T.~Lukasiewicz, Y.~Wu, J.~B. Tenenbaum, W.~Hart, T.~Gowers, W.~Li, A.~Weller, and M.~Jamnik.
\newblock Evaluating language models for mathematics through interactions.
\newblock \emph{Proceedings of the National Academy of Sciences}, 121\penalty0 (24):\penalty0 e2318124121, June 2024{\natexlab{a}}.
\newblock \doi{10.1073/pnas.2318124121}.
\newblock URL \url{https://www.pnas.org/doi/abs/10.1073/pnas.2318124121}.
\newblock Publisher: Proceedings of the National Academy of Sciences.

\bibitem[Collins et~al.(2024{\natexlab{b}})Collins, Sucholutsky, Bhatt, Chandra, Wong, Lee, Zhang, Zhi-Xuan, Ho, Mansinghka, Weller, Tenenbaum, and Griffiths]{collins_building_2024}
K.~M. Collins, I.~Sucholutsky, U.~Bhatt, K.~Chandra, L.~Wong, M.~Lee, C.~E. Zhang, T.~Zhi-Xuan, M.~Ho, V.~Mansinghka, A.~Weller, J.~B. Tenenbaum, and T.~L. Griffiths.
\newblock Building machines that learn and think with people.
\newblock \emph{Nature Human Behaviour}, 8\penalty0 (10):\penalty0 1851--1863, Oct. 2024{\natexlab{b}}.
\newblock ISSN 2397-3374.
\newblock \doi{10.1038/s41562-024-01991-9}.
\newblock URL \url{https://www.nature.com/articles/s41562-024-01991-9}.
\newblock Publisher: Nature Publishing Group.

\bibitem[Crockett(2025)]{crockett_modern_2025}
M.~J. Crockett.
\newblock Modern maxims for an {AI} oracle.
\newblock \emph{Nature Machine Intelligence}, 7\penalty0 (1):\penalty0 4--5, Jan. 2025.
\newblock ISSN 2522-5839.
\newblock \doi{10.1038/s42256-024-00970-z}.
\newblock URL \url{https://www.nature.com/articles/s42256-024-00970-z}.
\newblock Publisher: Nature Publishing Group.

\bibitem[Fischbach(2024)]{fischbach_problem_2024}
M.~A. Fischbach.
\newblock Problem choice and decision trees in science and engineering.
\newblock \emph{Cell}, 187\penalty0 (8):\penalty0 1828--1833, Apr. 2024.
\newblock ISSN 1097-4172.
\newblock \doi{10.1016/j.cell.2024.03.012}.

\bibitem[Gottweis et~al.(2025)Gottweis, Weng, Daryin, Tu, Palepu, Sirkovic, Myaskovsky, Weissenberger, Rong, Tanno, Saab, Popovici, Blum, Zhang, Chou, Hassidim, Gokturk, Vahdat, Kohli, Matias, Carroll, Kulkarni, Tomasev, Guan, Dhillon, Vaishnav, Lee, Costa, Penadés, Peltz, Xu, Pawlosky, Karthikesalingam, and Natarajan]{gottweis_towards_2025}
J.~Gottweis, W.-H. Weng, A.~Daryin, T.~Tu, A.~Palepu, P.~Sirkovic, A.~Myaskovsky, F.~Weissenberger, K.~Rong, R.~Tanno, K.~Saab, D.~Popovici, J.~Blum, F.~Zhang, K.~Chou, A.~Hassidim, B.~Gokturk, A.~Vahdat, P.~Kohli, Y.~Matias, A.~Carroll, K.~Kulkarni, N.~Tomasev, Y.~Guan, V.~Dhillon, E.~D. Vaishnav, B.~Lee, T.~R.~D. Costa, J.~R. Penadés, G.~Peltz, Y.~Xu, A.~Pawlosky, A.~Karthikesalingam, and V.~Natarajan.
\newblock Towards an {AI} co-scientist, Feb. 2025.
\newblock URL \url{http://arxiv.org/abs/2502.18864}.
\newblock arXiv:2502.18864 [cs].

\bibitem[Lombrozo(2024)]{lombrozo_learning_2024}
T.~Lombrozo.
\newblock Learning by thinking in natural and artificial minds.
\newblock \emph{Trends in Cognitive Sciences}, 28\penalty0 (11):\penalty0 1011--1022, Nov. 2024.
\newblock ISSN 1879-307X.
\newblock \doi{10.1016/j.tics.2024.07.007}.

\bibitem[Messeri and Crockett(2024)]{messeri_artificial_2024}
L.~Messeri and M.~J. Crockett.
\newblock Artificial intelligence and illusions of understanding in scientific research.
\newblock \emph{Nature}, 627\penalty0 (8002):\penalty0 49--58, Mar. 2024.
\newblock ISSN 1476-4687.
\newblock \doi{10.1038/s41586-024-07146-0}.
\newblock URL \url{https://www.nature.com/articles/s41586-024-07146-0}.
\newblock Publisher: Nature Publishing Group.

\bibitem[Oktar et~al.(2024)Oktar, Sucholutsky, Lombrozo, and Griffiths]{oktar_dimensions_2024}
K.~Oktar, I.~Sucholutsky, T.~Lombrozo, and T.~L. Griffiths.
\newblock Dimensions of disagreement: {Divergence} and misalignment in cognitive science and artificial intelligence.
\newblock \emph{Decision}, 11\penalty0 (4):\penalty0 511--522, 2024.
\newblock ISSN 2325-9973.
\newblock \doi{10.1037/dec0000244}.
\newblock Place: US Publisher: Educational Publishing Foundation.

\bibitem[Plato()]{plato_phaedrus_nodate}
Plato.
\newblock \emph{Phaedrus}.
\newblock Penguin Classics.
\newblock ISBN 978-0-14-044974-7.

\bibitem[Riedl et~al.(2021)Riedl, Kim, Gupta, Malone, and Woolley]{riedl_quantifying_2021}
C.~Riedl, Y.~J. Kim, P.~Gupta, T.~W. Malone, and A.~W. Woolley.
\newblock Quantifying collective intelligence in human groups.
\newblock \emph{Proceedings of the National Academy of Sciences}, 118\penalty0 (21):\penalty0 e2005737118, May 2021.
\newblock \doi{10.1073/pnas.2005737118}.
\newblock URL \url{https://www.pnas.org/doi/abs/10.1073/pnas.2005737118}.
\newblock Publisher: Proceedings of the National Academy of Sciences.

\bibitem[Shavit et~al.(2023)Shavit, Agarwal, Brundage, Adler, O’Keefe, Campbell, Lee, Mishkin, Eloundou, Hickey, et~al.]{shavit2023practices}
Y.~Shavit, S.~Agarwal, M.~Brundage, S.~Adler, C.~O’Keefe, R.~Campbell, T.~Lee, P.~Mishkin, T.~Eloundou, A.~Hickey, et~al.
\newblock Practices for governing agentic ai systems.
\newblock \emph{Research Paper, OpenAI}, 2023.

\bibitem[Youn et~al.(2015)Youn, Strumsky, Bettencourt, and Lobo]{youn_invention_2015}
H.~Youn, D.~Strumsky, L.~M.~A. Bettencourt, and J.~Lobo.
\newblock Invention as a combinatorial process: evidence from {US} patents.
\newblock \emph{Journal of The Royal Society Interface}, 12\penalty0 (106):\penalty0 20150272, May 2015.
\newblock \doi{10.1098/rsif.2015.0272}.
\newblock URL \url{https://royalsocietypublishing.org/doi/10.1098/rsif.2015.0272}.
\newblock Publisher: Royal Society.

\bibitem[Zhou et~al.(2024)Zhou, Liu, Jin, Nie, and Dou]{zhou_metacognitive_2024}
Y.~Zhou, Z.~Liu, J.~Jin, J.-Y. Nie, and Z.~Dou.
\newblock Metacognitive {Retrieval}-{Augmented} {Large} {Language} {Models}.
\newblock In \emph{Proceedings of the {ACM} {Web} {Conference} 2024}, {WWW} '24, pages 1453--1463, New York, NY, USA, May 2024. Association for Computing Machinery.
\newblock ISBN 9798400701719.
\newblock \doi{10.1145/3589334.3645481}.
\newblock URL \url{https://dl.acm.org/doi/10.1145/3589334.3645481}.

\end{thebibliography}
\bibliographystyle{abbrvnat}

%%%%%%%%%%%%%%%%%%%%%%%%%%%%%%%%%%%%%%%%%%%%%%%%%%%%%%%%%%%%%%%%%%%%%%%%%%%%%%%
%%%%%%%%%%%%%%%%%%%%%%%%%%%%%%%%%%%%%%%%%%%%%%%%%%%%%%%%%%%%%%%%%%%%%%%%%%%%%%%
% APPENDIX
%%%%%%%%%%%%%%%%%%%%%%%%%%%%%%%%%%%%%%%%%%%%%%%%%%%%%%%%%%%%%%%%%%%%%%%%%%%%%%%
%%%%%%%%%%%%%%%%%%%%%%%%%%%%%%%%%%%%%%%%%%%%%%%%%%%%%%%%%%%%%%%%%%%%%%%%%%%%%%%
\appendix

% \section{Appendix}

% \begin{figure*}[h!]
%  \begin{center}
%  \includegraphics[width=1.0\linewidth]{Submission/figures/elic_interface.png}
%   \caption{Example relabeling elicitation task shown to each crowdsourced worker.}
%     \label{fig:elicReLabel}
%  \end{center}
%  \end{figure*}
 
%  \begin{figure*}[h!]
%  \begin{center}
%  \includegraphics[width=1.0\linewidth]{Submission/figures/uncertain_elic.png}
%   \caption{Example soft label elicitation task shown to each crowdsourced worker.}
%     \label{fig:elicSoftLabels}
%  \end{center}
%  \end{figure*}

%%%%%%%%%%%%%%%%%%%%%%%%%%%%%%%%%%%%%%%%%%%%%%%%%%%%%%%%%%%%%%%%%%%%%%%%%%%%%%%
%%%%%%%%%%%%%%%%%%%%%%%%%%%%%%%%%%%%%%%%%%%%%%%%%%%%%%%%%%%%%%%%%%%%%%%%%%%%%%%

\end{document}